\documentclass[10pt,twocolumn,letterpaper]{article}

\usepackage[pagenumbers]{cvpr} 
\usepackage{multirow}
\definecolor{cvprblue}{rgb}{0.21,0.49,0.74}
\usepackage[pagebackref,breaklinks,colorlinks,allcolors=cvprblue]{hyperref}

\def\paperID{} 
\def\confName{CVPR}
\def\confYear{2026}

\title{
Mirage: One-Step Video Diffusion for Photorealistic and Coherent \\ Asset Editing in Driving Scenes}

\author{
    Shuyun Wang$^{1,2}$ \quad 
    Haiyang Sun$^2$\textsuperscript{\footnotemark[2]}\quad 
    Bing Wang$^2$ \quad
    Hangjun Ye$^2$\textsuperscript{\footnotemark[1]} \quad
    Xin Yu$^1$\textsuperscript{\footnotemark[1]}\\
    $^1$The University of Queensland \quad 
    $^2$Xiaomi EV \\
    {\tt\small \{shuyun.wang\}@uq.edu.au}
}

\begin{document}

\twocolumn[{
\maketitle
\vspace{-2.5em}
\begin{center}
    \centering
    \includegraphics[width=\textwidth]{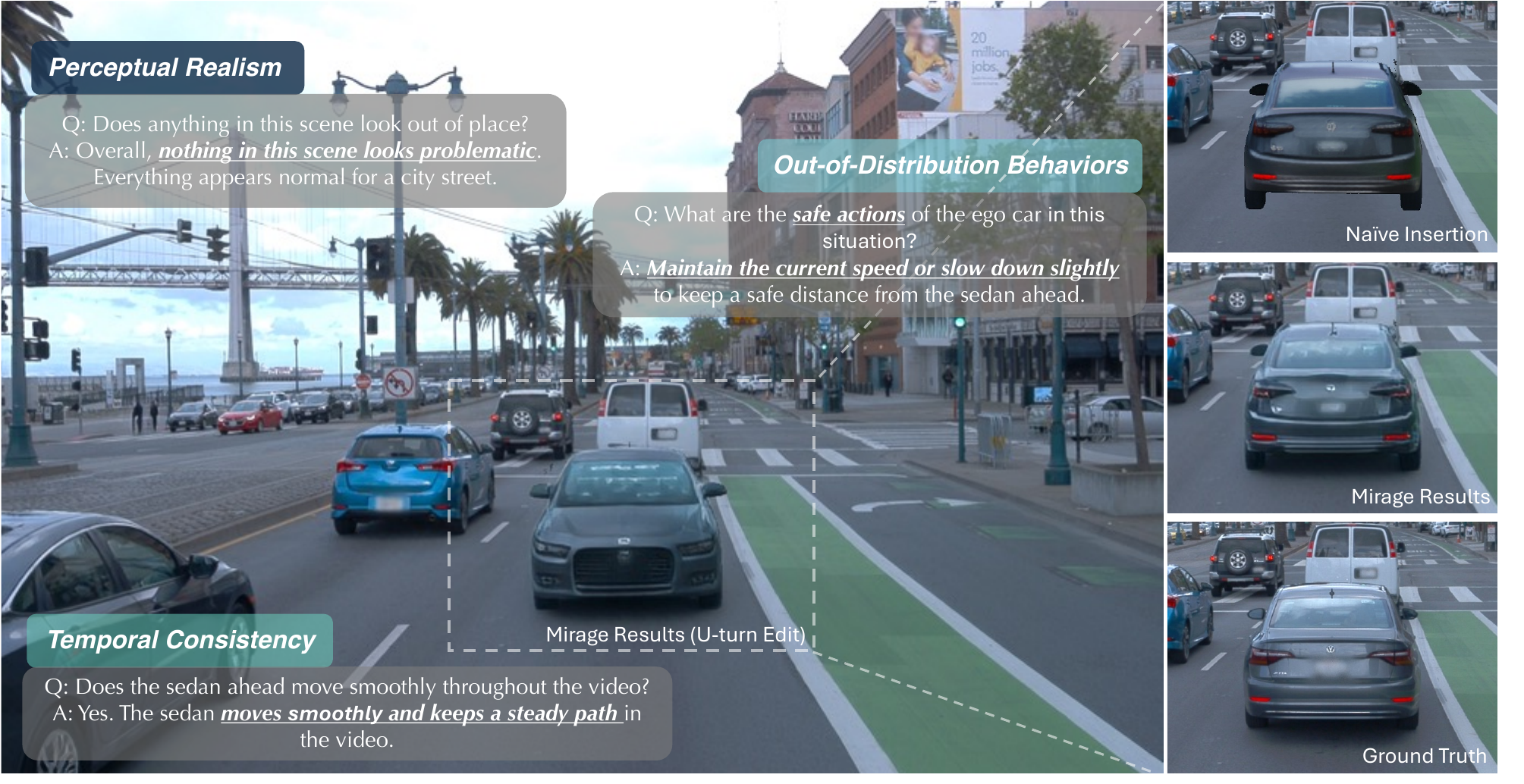}
    \vspace{-1.7em}
    \captionof{figure}{
    \textbf{Photorealistic results of Mirage.} GPT-4V visual question answering reveals three key properties: (i) perceptual realism, with no detected inconsistencies; (ii) behavioral out-of-distribution effects, which lead to incorrect reasoning; and (iii) temporal consistency, with full video results provided in the supplementary material. Right: Comparison of naïve insertion, Mirage results, and ground truth.
    }
    \label{fig:teaser}
\end{center}
}]

\setcounter{footnote}{0}
\footnotetext[2]{Project leader; %
\textsuperscript{*}Corresponding author
}

\begin{abstract}
Vision-centric autonomous driving systems rely on diverse and scalable training data to achieve robust performance.
While video object editing offers a promising path for data augmentation, existing methods often struggle to maintain both high visual fidelity and temporal coherence.
In this work, we propose \textbf{Mirage}, a one-step video diffusion model for photorealistic and coherent asset editing in driving scenes.
Mirage builds upon a text-to-video diffusion prior to ensure temporal consistency across frames.
However, 3D causal variational autoencoders often suffer from degraded spatial fidelity due to compression, and directly passing 3D encoder features to decoder layers breaks temporal causality. 
To address this, we inject temporally agnostic latents from a pretrained 2D encoder into the 3D decoder to restore detail while preserving causal structures.
Furthermore, because scene objects and inserted assets are optimized under different objectives, their Gaussians exhibit a distribution mismatch that leads to pose misalignment.
To mitigate this, we introduce a two-stage data alignment strategy combining coarse 3D alignment and fine 2D refinement, thereby improving alignment and providing cleaner supervision.
Extensive experiments demonstrate that Mirage achieves high realism and temporal consistency across diverse editing scenarios. 
Beyond asset editing, Mirage can also generalize to other video-to-video translation tasks, serving as a reliable baseline for future research. Our code is available at \url{https://github.com/wm-research/mirage}.
\end{abstract}
    
\section{Introduction}
\label{sec:intro}

Vision-centric autonomous driving (AD) systems rely on large-scale and diverse training data to build reliable perception models.
Public benchmarks such as nuScenes~\cite{nuscenes} and Waymo Open Dataset~\cite{Waymo} have played an important role by providing extensive real-world driving videos.
However, these datasets show a long-tail distribution.
Common driving scenarios appear frequently.
Safety-critical corner cases, such as unexpected insertions and rare maneuvers, are highly underrepresented.
This imbalance leads to reduced generalization in challenging environments and weakens performance in safety-critical situations.
Collecting such corner case data is difficult, expensive, and risky, but it is still necessary for developing robust autonomous driving systems.


To augment rare driving scenarios, generative methods~\cite{vista, panacea,drivedreamer} have been proposed to synthesize data from scene layouts like BEV maps or text prompts.
These scene-level approaches lack fine-grained control over object pose and appearance.
This limitation reduces their effectiveness in generating precise corner cases.
Object-centric video editing~\cite{dipir, g2editor, driveeditor, genmm} enables better control through 2D inpainting guided by 3D boxes and reference images.
Since it operates only in the 2D image domain without reasoning about 3D structures, it struggles to maintain accurate geometry and multi-view consistency.
Recently, R3D2~\cite{r3d2} introduced a more promising approach by inserting 3D assets into reconstructed scenes and harmonizing the outputs using image diffusion.
This hybrid pipeline offers more accurate pose control than inpainting-based methods.
However, mismatches between 3D assets and scene Gaussians can still lead to geometric misalignments.
In addition, performing harmonization independently on each frame often results in temporal inconsistency.

To address these issues, we propose \textbf{Mirage}, a one-step video diffusion model that integrates external 3D assets into driving scenes with photorealistic and coherent rendering effects.
Built upon a powerful pretrained text-to-video foundation, Mirage leverages its inherent temporal priors to ensure cross-frame consistency without auxiliary temporal modules.
However, adapting one-step image diffusion to the video domain is non-trivial.
Latent compression in variational autoencoders (VAEs) often leads to the loss of high-frequency details. 
Previous 2D VAE-based methods~\cite{difix3d+, img2imgturbo, r3d2} pass intermediate encoder features to the decoder to preserve spatial details.
However, directly passing 3D encoder features introduces temporal leakage due to the asymmetric nature of encoder and decoder activations across time.
To resolve this, we introduce a \textit{temporally agnostic latent injection} strategy.
We observe that decoder latents at later upsampling stages exhibit a one-to-one correspondence with output frames.
Building on this insight, we inject temporally agnostic, high-frequency latents extracted from a pretrained 2D encoder into corresponding upsampling layers. 
This enables the injection of high-frequency details without interfering with the causal temporal structure of the 3D decoder.
Since these injected 2D latents reside in a different representational space, we introduce a cross-modal fusion block prior to the upsampling block.
Additionally, we train a dedicated 3D Low-Rank adaptation (LoRA) on the 3D decoder to better align the injected 2D latents with the temporal latent space of the 3D decoder.
This design preserves temporal isolation while enhancing spatial fidelity, achieving both photorealism and frame-to-frame consistency in the output videos.

Moreover, we observe that in previous asset reinsertion pipelines, the distribution gap between scene Gaussians and 3D assets often leads to inaccurate pose alignment and produces noisy supervisory signals.
To bridge this gap, we introduce a two-stage alignment strategy.
We first align the asset’s Gaussian ellipsoids to the scene object by optimizing its center, orientation, and scale.
Subsequently, we refine the alignment in 2D by calibrating the rendered asset’s bounding box against the ground-truth bounding box, ensuring tighter image-space consistency.
This procedure significantly improves alignment accuracy and yields cleaner, more stable supervision.
Following this pipeline, we construct MirageDrive, a dataset of 3,550 video clips (2,840 training clips with 710 validation clips) with precisely aligned asset reinsertions.
Equipped with our diffusion model and the MirageDrive dataset, our Mirage achieves impressive performance.

Extensive experiments demonstrate that Mirage achieves superior visual realism and temporal coherence, outperforming existing methods across both qualitative assessments and quantitative metrics.
Even under challenging editing conditions, such as repositioning a vehicle with a 180-degree direction reversal, Mirage produces outputs that are visually indistinguishable from real videos, as shown in Fig.~\ref{fig:teaser}.
Moreover, our framework also shows strong potential for general video-to-video translation tasks such as relighting or restoration.

In summary, our key contributions are as follows:
\begin{itemize}
\item We present Mirage, a generalizable one-step video diffusion model that generates realistic and temporally coherent rendering effects for 3D asset insertion.
\item We introduce a temporally-agnostic latent injection strategy that enhances spatial fidelity while strictly preserving temporal causality in causal 3D VAEs.
\item We develop a two-stage alignment strategy and create MirageDrive, a high-quality dataset of 3,550 clips that provides precise alignments and clean supervisory signals.
\end{itemize}
\section{Related Work}
\label{sec:relatedwork}

\subsection{One-Step Diffusion Model}
A prevalent strategy for accelerating diffusion models is to reduce the number of sampling steps. Common technical routes include rectified flow~\cite{rectifiedflow,instaflow}, score distillation~\cite{zhou2024score,wang2023prolificdreamer,yin2024one}, and adversarial training~\cite{lin2025diffusion,zhang2024sf}. Building on these advances, Image-to-Image Turbo~\cite{img2imgturbo} demonstrates how adversarial fine-tuning enables photorealistic image translation in a single step. In 3D scene rendering, DIFIX3D+~\cite{difix3d+} employs variational score distillation to enhance novel view synthesis quality while maintaining 3D consistency. Extending such acceleration techniques to the video domain, however, introduces the critical challenge of maintaining temporal coherence. The recently proposed DOVE~\cite{dove} addresses this by fine-tuning a pre-trained text-to-video model for one-step video super-resolution. Although it achieves improved coherence, its reliance on a 3D VAE for latent representation introduces noticeable compression artifacts, ultimately constraining the visual quality of the output.

\subsection{Video Editing in Autonomous Driving}
The evolution of video editing, particularly with diffusion models~\cite{videop2p,pix2video,mgdm}, has significantly advanced object-level manipulations. However, such methods often lack 3D spatial reasoning, limiting their direct applicability in autonomous driving scenarios. To address this gap, several driving-specific editing approaches have been developed. Methods like GenMM~\cite{genmm} adopt an inpainting framework for object insertion, while DriveEditor~\cite{driveeditor} introduces a unified system supporting multiple editing tasks. Similarly, $G^2$Editor leverages a 3D Gaussian representation as a dense prior to achieve more geometrically precise editing. Despite these contributions, these methods still face challenges in precise pose control and visual realism. In a parallel line of research, 3D Gaussian Splatting (3DGS)-based methods such as OmniRe~\cite{omnire} and SplatAD~\cite{splatad} have emerged, enabling object-level editing through Gaussian node manipulation. While these techniques provide more accurate 3D spatial control, they often struggle with inconsistent lighting conditions and output realism.

\subsection{Asset Insertion and Relighting}
Seamless asset insertion and relighting constitute a fundamental challenge in video editing. 
Traditional Physically Based Rendering (PBR) methods~\cite{physically} provide principled solutions but require accurate material properties and detailed geometry. 
Diffusion-based methods like DiPIR~\cite{dipir} combine generative models with inverse rendering, yet still necessitate asset material knowledge.
Recent generative relighting approaches have explored more flexible paradigms. 
IC-Light~\cite{iclight} adapts 2D object appearance to new scenes but often neglects reciprocal scene modifications like shadow casting.
Similarly, DiffHarmony++~\cite{diffharmony} addresses image harmonization through latent diffusion, yet primarily focuses on adapting foreground appearance without explicit consideration for bidirectional object-scene lighting interaction.
In the context of autonomous driving, R3D2~\cite{r3d2} employs a one-step diffusion model to generate realistic rendering effects for asset insertion. However, its single-image processing paradigm lacks explicit temporal modeling, limiting its ability to maintain consistency across video sequences.
\section{Method}
\begin{figure*}[t]
  \centering
  \vspace{-1em}
   \includegraphics[width=\textwidth]{fig/backbone.pdf}
    \caption{\textbf{Overview of the framework and training flow of Mirage.} Given a naïve insertion video, Mirage encodes the input using both a 2D VAE encoder and a causal 3D VAE encoder. The 2D intermediate latents are fused with 3D latents through cross-modal fusion blocks. A diffusion transformer, equipped with 2D LoRA adapters, refines the NI latent toward the ground-truth (GT) latent. The 3D VAE decoder then reconstructs the edited video using two sets of LoRA modules: Reconstruction-LoRA for VAE adaptation and Harmonization-LoRA for harmonization fine-tuning.}
   \label{fig:backbone}
     \vspace{-1em}
\end{figure*}

We decouple the task of photorealistic and coherent asset editing into two subtasks: (1) accurate scene–asset interaction, and (2) harmonization with the surrounding scene.
Precise editing is ensured by our two-stage alignment strategy, which we use to curate a high-quality dataset with clean and consistently aligned 3D insertions for reliable one-step diffusion supervision.
To further enhance visual realism, we adopt a two-stage training strategy—consisting of a VAE adaptation stage and a harmonization training stage that enables photorealistic rendering of edited outputs.
\subsection{Preliminary}
Diffusion models are a class of generative models that synthesize data by reversing a forward noising process.
In our setting, generation operates in latent space, where a clean latent variable $z_0$ is gradually corrupted into a noisy version $z_t$ through a fixed Gaussian noise schedule.
Following standard practice, we sample $z_t$ using the closed-form expression: $
z_t = \sqrt{\bar{\alpha}_t} z_0 + \sqrt{1 - \bar{\alpha}_t}\epsilon$,
where $\epsilon \sim \mathcal{N}(0, I)$ is standard Gaussian noise, and $\bar{\alpha}_t$ denotes the cumulative signal preservation factor at diffusion timestep $t$.
The reverse process is learned by a neural network $\epsilon_\theta$, which is trained to predict the added noise $\epsilon$ given the optional conditioning inputs $c$ (e.g., text, masks, or images). The model is optimized using the noise prediction loss:
\begin{equation}
\label{preliminaries1}
\mathcal{L} = \mathbb{E}_{z_0, t, \epsilon}[\|\ \epsilon_\theta(z_t, t, c) - \epsilon \|^2_2].
\end{equation}

While traditional diffusion models generate outputs through multiple denoising steps, recent approaches have explored one-step diffusion strategies for faster inference.
These methods directly predict the clean latent $z_0$ from a single noisy input $z_t$ at a fixed timestep $t$, often by modifying the network architecture or adjusting the training scheme to strengthen the denoising prior.

\subsection{Overall Framework}
We construct our one-step video diffusion model, \textbf{Mirage}, based on a powerful pretrained text-to-video model, CogVideoX~\cite{cogvideox}. 
CogVideoX employs a causal 3D VAE to encode videos into a spatiotemporal latent space, and a Transformer denoiser $\epsilon_\theta$ to perform latent-space denoising. 
Thanks to its strong temporal priors, Mirage ensures temporal coherence without requiring auxiliary motion-specific modules such as temporal attention or optical flow.

The overall architecture of Mirage is illustrated in Fig.~\ref{fig:backbone}.
At inference time, given a naively insertion video $x_{NI}$, we first encode it into spatiotemporal latent codes $z_{NI}$ using the 3D VAE encoder.
In parallel, we also process $x_{NI}$ through a pretrained 2D VAE encoder and cache intermediate latents from selected layers for later injection. 
Following prior works~\cite{r3d2,difix3d+}, we treat $z_{NI}$ as the noised latent $z_t$ at a fixed timestep $t$ and perform one-step denoising using $\epsilon_\theta$, producing the denoised latent $z_{DR}$.
Since we only execute a single denoising step, we set a smaller timestep $t{=}199$ to balance denoising strength and structural preservation.
Finally, $z_{DR}$ is decoded by the 3D VAE decoder to generate the edited output video $x_{DR}$.
During decoding, we inject the cached high-frequency latents from the 2D VAE into selected layers of the 3D decoder using our temporally agnostic latent injection strategy, which enhances spatial fidelity while maintaining temporal consistency.

\subsubsection{VAE Adaption Stage}
\paragraph{Motivation.}
A core limitation of latent diffusion models lies in spatial detail loss due to heavy compression introduced by the VAEs. The
2D encoder typically downsamples inputs by a factor of $8\times$, resulting in loss of fine details.
This issue becomes even more severe in 3D VAEs, where causal 3D VAEs~\cite{cogvideox, magvit2, wan} often additionally apply $4\times$ temporal downsampling to improve efficiency.
While 2D models mitigate this degradation by introducing skip connections between encoder and decoder layers~\cite{difix3d+, img2imgturbo}, naively applying the same strategy to 3D VAEs introduces severe temporal artifacts.
Due to the causal nature of 3D encoders $E^{3D}$ and decoders $D^{3D}$, skip connections disrupt the temporal distribution of features, resulting in ghosting and flickering across frames.

\vspace{-1em}
\paragraph{Temporally Agnostic Latent Injection.}
To overcome this limitation, we introduce a \textit{temporally agnostic latent injection} strategy that enhances spatial fidelity while preserving temporal causality.
Our key observation is that in the last two upsampling blocks of $D^{3D}$, each latent slice corresponds one-to-one with the output frames.
This structure allows safe injection of static features without disrupting causal temporal modeling.

We first encode the naively inserted video $x_{NI}$ using a pretrained 2D VAE encoder $E^{2D}$ to extract high-frequency intermediate latents ${z}^{2D}_{{mid}}$.
These latents are reshaped from the batch dimension to the temporal dimension and injected into $D^{3D}$ via Cross-Modal Fusion Blocks (CMFBs).
Each CMFB fuses the 2D and 3D intermediate latents through concatenation and merging operations, enriching the decoder representation with detailed spatial cues while maintaining temporal isolation.

However, since ${z}^{2D}_{{mid}}$ and ${z}^{3D}_{{mid}}$ lie in different representational spaces, we introduce trainable 3D causal LoRA adapters (denoted as Reconstruction-LoRA) into $D^{3D}$.
These lightweight adapters modulate the convolutional weights, enabling the decoder to interpret injected cross-modal features effectively with minimal finetuning.

\vspace{-1em}
\paragraph{Training Objective.}
During this stage, we decouple training from the diffusion process and directly supervise the decoder to reconstruct a high-quality output video.
Although the latent input ideally should be sampled from the diffusion process, we empirically find that using the encoded latent of the ground-truth video $x_{GT}$ provides a close approximation when the denoiser is well-trained.
The overall training process is formulated as:
\begin{equation}
{x}_{RO} = D^{3D}[E^{3D}(x_{GT}), z^{2D}_{mid}],
\end{equation}
where $z^{2D}_{mid}$ denotes the injected static latents extracted from $E^{2D}(x_{NI})$.
As illustrated in Fig.~\ref{fig:backbone}, we minimize the reconstruction error between the output and the GT using a combination of mean squared error (MSE) loss and perceptual loss ($\mathcal{L}_{lpips}$~\cite{lpips}):
\begin{equation}
\mathcal{L}_{vae} = \mathcal{L}_{mse}(x_{RO} , x_{GT}) + \lambda_1 \mathcal{L}_{lpips}(x_{RO} , x_{GT}),
\end{equation}
where $\lambda_1$ is a scalar weight for the perceptual loss. This training strategy substantially enhances reconstruction fidelity and texture realism in the final outputs.

\begin{figure*}[t]
  \centering
   \includegraphics[width=\linewidth]{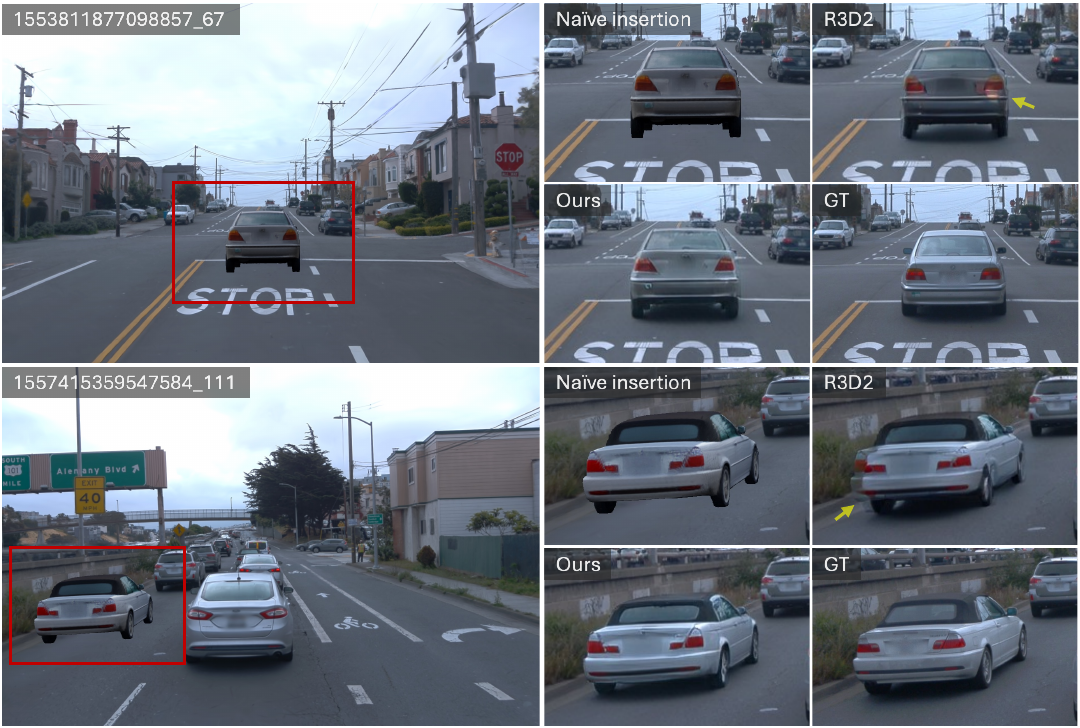}
    \caption{\textbf{Qualitative comparison of asset insertion results}. Naïve insertion produces visible artifacts and inconsistent shading, while R3D2 partially improves realism but still shows lighting and geometry mismatches. Mirage generates photorealistic, temporally consistent vehicles that closely match ground-truth appearance.
    }
   \label{fig:exp_1}
\end{figure*}

\begin{figure*}[t]
  \centering
   \includegraphics[width=\linewidth]{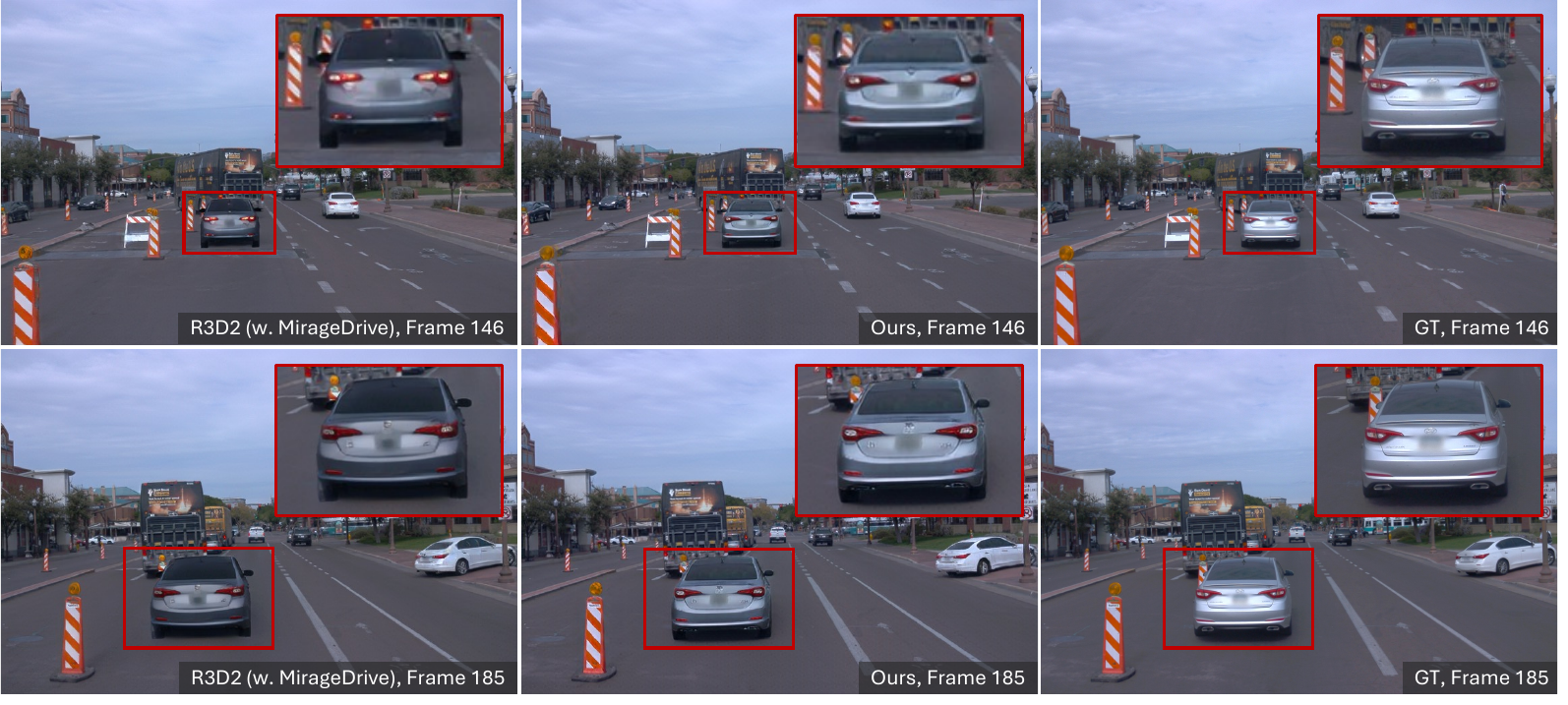}
    \caption{\textbf{Vehicle identity preservation comparison.} We retrain R3D2 on the MirageDrive dataset to ensure consistent geometry before evaluating temporal behavior. Even under this controlled setting, R3D2 shows clear appearance drift across distant frames. In contrast, Mirage preserves stable shape, reflectance, and vehicle identity, remaining closely aligned with the ground truth.
    }
   \label{fig:exp_2}
\end{figure*}

\subsubsection{Harmonization Training Stage}
After the VAE adaptation stage, Mirage effectively reconstructs videos from naively inserted scenes with improved spatial fidelity. However, subtle rendering cues such as shadows, specular reflections, and material consistency are still not fully recovered. These effects are crucial for photorealistic harmonization in image space.

To further enhance appearance quality and realism, we fine-tune Mirage in pixel space while freezing all previously introduced parameters, including the cross-modal fusion blocks and the Reconstruction-LoRA.
We introduce two dedicated modules for detail refinement.
First, we inject trainable 2D LoRA layers into the pretrained Transformer denoiser $\epsilon_\theta$, enabling the model to adapt its attention blocks for harmonization.
Second, we incorporate a causal 3D LoRA (Harmonization-LoRA) into the 3D VAE decoder. These layers structurally extend 2D LoRA by replacing 2D convolutions with causal 3D convolutions, ensuring each output frame depends only on past context and preserving the temporal causal structure.

\vspace{-1em}
\paragraph{Training Objective.}
The model is trained to enhance visual realism in the diffusion render $x_{DR}$ given a naïvely inserted input $x_{NI}$.
In this stage, we use LPIPS loss to preserve perceptual structure and a Gram-matrix loss~\cite{gram} to match texture statistics:
\begin{equation}
\mathcal{L}_{harmon.} = \mathcal{L}_{lpips}(x_{DR}, x_{GT}) + \lambda_2 \mathcal{L}_{gram}(x_{DR}, x_{GT}),
\end{equation}
where $\lambda_2$ is a scalar weight for the Gram loss. The $\mathcal{L}_{lpips}$ enforces similarity in deep feature space, while $\mathcal{L}_{gram}$ captures second-order texture statistics by computing Gram matrices over VGG-16 feature activations.

\subsection{Data Processing Pipeline}
To support the supervised training of Mirage, we construct a high-quality dataset of driving videos with precisely aligned 3D asset reinsertions. A key challenge in building such a dataset lies in reducing supervision noise caused by pose mismatches between inserted assets and scene objects.

\vspace{-1em}
\paragraph{Motivation.}
In contrast to prior work such as R3D3~\cite{r3d2}, which directly replaces object Gaussians with generated assets, our setting reveals a notable distribution mismatch: the 3D Gaussians used in scene reconstruction and those representing inserted assets are optimized under different objectives and originate from distinct data distributions.
As a result, naïvely aligning them based solely on position often leads to misalignment, which introduces noisy residuals into the training supervision. Since Mirage decouples geometry editing from appearance harmonization, such misalignment severely hinders the latter’s ability to learn meaningful render effects and produces inconsistent outputs.

\vspace{-1em}
\paragraph{Two-Stage Alignment Strategy.}
To address the misalignment caused by Gaussian distribution mismatch, we introduce a two-stage alignment strategy that ensures high-quality scene–asset interaction through both 3D and 2D alignment.

We begin with a coarse 3D alignment step, where the inserted asset's Gaussian set is matched to that of the original scene object. This process estimates a rigid transformation (translation, rotation, and scaling) that places the asset approximately in the same world-space location as the original object. After rendering the video with the coarsely aligned asset, we compute 2D bounding boxes of the asset across all frames and compare them to the original object’s annotated boxes. We then estimate a global affine transformation (displacement and scaling) by averaging the bounding-box differences over the entire video. This transformation is uniformly applied to all frames to ensure tight spatial alignment in image space, correcting residual misalignment from 3D matching and improving frame-level consistency.

\vspace{-1em}
\paragraph{Data Pair Curation.}
Given this two-stage alignment, we generate training pairs $(x_{NI}, x_{GT})$, where $x_{NI}$ is the video rendered from the scene with aligned 3D asset insertions and $x_{GT}$ is the original driving video. The former captures correct object placement but lacks realistic visual effects (e.g., lighting, shadows), while the latter provides the photorealistic reference. These pairs enable the harmonization training stage of Mirage to focus on learning meaningful render effects without being confounded by pose misalignment.

We apply the above data curation pipeline to the Waymo Open Dataset~\cite{Waymo}. In total, we construct a dataset of 3,550 training video clips with carefully aligned 3D insertions across diverse object categories, viewpoints, and lighting conditions. This forms the foundation for both VAE adaptation and harmonization training stages.
\begin{table*}[t]
    \centering
    \caption{Quantitative comparison on our evaluation set. $\uparrow$ indicates higher is better; $\downarrow$ indicates lower is better.}
        \vspace{-0.5em}
    \label{tab:quantitative}
        \begin{tabular}{l | c c c c c c | c}
        \toprule
            Method & PSNR$\uparrow$ & SSIM$\uparrow$ & LPIPS$\downarrow$ & DISTS$\downarrow$ & $\text{E}_{warp}^*\downarrow$ & vFID$\downarrow$ & FPS$\uparrow$ \\
            \midrule
            \multicolumn{8}{c}{\textbf{Actor-centric crops}} \\
            \midrule
            Naive Ins. & 14.93 & 0.307 & 0.209 & 0.236 & \textbf{0.801} & 0.506 & -- \\
            \midrule
            R3D2 & 16.18 & 0.356 & 0.188 & 0.207 & 1.119 & \textbf{0.442} & -- \\
            Mirage & \textbf{17.43} & \textbf{0.390} & \textbf{0.179} & \textbf{0.183} & 0.874 & 0.448 & -- \\
            \midrule
            \multicolumn{8}{c}{\textbf{Full-resolution}} \\
            \midrule
            Naive Ins. & 27.56 & \textbf{0.921} & 0.112 & 0.091 & \textbf{7.423} & 0.244 & -- \\
            \midrule
            R3D2 & 27.61 & 0.904 & 0.072 & 0.055 & 11.320 & 0.206 & 4.07 \\
            Mirage & \textbf{28.60} & 0.914 & \textbf{0.069} & \textbf{0.054} & 9.955 & \textbf{0.178} & \textbf{4.54} \\
            \bottomrule
        \end{tabular}
        \vspace{-0.5em}
\end{table*}

\section{Experiments}

\subsection{Experimental Settings}
\paragraph{Implementation Details.}
Our Mirage framework is built upon the text-to-video model CogVideoX1.5~\cite{cogvideox}.
To reduce inference overhead, we use a fixed prompt, “remove degradation,” and pre-encode its text embedding.
Both training stages are trained on 8 NVIDIA H200 GPUs with a total batch size of 8, using the AdamW optimizer~\cite{adamw} with $\beta_1 = 0.9$ and $\beta_2 = 0.999$.
The training data consist of 9-frame clips at a resolution of $512 \times 768$.
Each stage runs for 10,000 iterations with a learning rate of $1 \times 10^{-4}$.
The loss weights $\lambda_1$ and $\lambda_2$ are both set to 0.1.
In the harmonization training stage, we activate the Gram loss after 2,000 steps and apply a constant warm-up schedule for the first 500 steps.
To construct our MirageDrive dataset, we employ Trellis~\cite{trellis} as the 3D asset generation framework.
Each sequence is reconstructed into a virtual environment using the neural reconstruction method SplatAD~\cite{splatad}. For the 2D latent space, we adopt AutoEncoderKL as our 2D VAE.

\vspace{-1em}
\paragraph{Evaluation Metrics.}
We adopt multiple evaluation metrics to assess model performance. For spatial fidelity, we report PSNR and SSIM~\cite{ssim}. 
For perceptual quality, we use LPIPS~\cite{lpips} and DISTS~\cite{dists}. 
To assess temporal consistency, we compute the average flow warping error $E^*_{warp}$~\cite{ewarp, dloral}, where each frame is warped using optical flow from ground truth. 
To evaluate overall realism, we compute video Fréchet Inception Distance (vFID)~\cite{vfid} between generated videos and their ground-truth counterparts. 
These metrics jointly assess reconstruction accuracy, perceptual quality, and temporal stability.

\subsection{Experimental Results}

\paragraph{Quantitative Comparison.}
In Tab.~\ref{tab:quantitative}, we report quantitative results on our validation set, covering both actor-centric crops and full-resolution images. The actor-centric setting provides a focused assessment of local editing fidelity near the inserted asset, while the full-resolution evaluation reflects global consistency and overall scene realism. As shown, naïve insertion performs poorly especially under actor-centric evaluation, indicating strong inconsistencies between the inserted asset and the surrounding context. R3D2 improves over naïve insertion in most metrics. Mirage further improves the results across nearly all metrics. On actor-centric crops, Mirage achieves lower LPIPS and DISTS compared to R3D2, demonstrating better alignment with human perceptual preferences. The reduced $\text{E}_{warp}^*$ and vFID suggests that Mirage generates more stable temporal transitions around edited regions. On full-resolution images, Mirage consistently surpasses both baselines in all metrics, showing that it preserves global scene structure while producing visually coherent and photorealistic edits. Moreover, Mirage achieves slightly higher FPS than R3D2, indicating that the proposed architecture improves realism without introducing additional computational overhead.

\vspace{-1em}
\paragraph{Qualitative Comparison.}
Fig.~\ref{fig:exp_1} presents a qualitative comparison between naïve insertion, R3D2, and our Mirage across challenging driving scenes. Naïve insertion often produces clear visual discontinuities between the inserted vehicle and the surrounding environment, including missing shadows, inconsistent illumination, and noticeable boundary artifacts. R3D2 reduces some of these inconsistencies but still exhibits shading mismatches, blurred reflections, and incomplete interaction with scene geometry.

In contrast, Mirage generates photorealistic results. Our method accurately infers scene lighting and produces consistent shadows beneath the inserted vehicles. In addition, the inserted assets blend naturally into the road structure and scene layout, yielding results that closely match real imagery. These examples highlight Mirage’s ability to maintain both high spatial fidelity and strong scene-level realism.

\begin{table}[t]
    \centering
    \caption{Ablation study on VAE adaptation strategies. (“Skip Con.” denotes skip connections; “2D Inj.” denotes injecting 2D encoder latents.)}
    \label{tab:vae_ablation}
    \resizebox{\linewidth}{!}{
        \begin{tabular}{l | c c c c}
        \toprule 
        Method & PSNR$\uparrow$ & LPIPS$\downarrow$ & vFID$\downarrow$ \\
        \midrule \midrule
        2D~VAE & 30.30 & 0.040 & 0.112 \\
        2D~VAE (w. Skip~Con.) & 32.70 & 0.043 & 0.106 \\
        3D~VAE & 33.98 & 0.041 & 0.080 \\
        3D~VAE (w. 2D~Inj.) & \textbf{35.07} & \textbf{0.039} & \textbf{0.079} \\
        \bottomrule
    \end{tabular}
    }
\end{table}

\begin{table}[t]
    \centering
    \caption{Comparison of R3D2 with and without MirageDrive.}
    \label{tab:dataset_ablation}
    \resizebox{\linewidth}{!}{
        \begin{tabular}{l | c c c c}
        \toprule 
        Method & PSNR$\uparrow$ & LPIPS$\downarrow$ & vFID$\downarrow$ \\
        \midrule \midrule
        R3D2 & 27.61 & 0.072 & 0.206 \\
        R3D2 (w. MirageDrive) & \textbf{28.31} & 0.072 & \textbf{0.193} \\
        \bottomrule
    \end{tabular}
    }
\end{table}

\begin{table}[t]
    \centering
    \caption{Ablation study on different training configurations.}
    \label{tab:training_ablation}
    \resizebox{\linewidth}{!}{
    \begin{tabular}{c c c | c c c}
        \toprule 
        \multicolumn{3}{c|}{Configuration} & \multirow{2}{*}{PSNR$\uparrow$} & \multirow{2}{*}{LPIPS$\downarrow$} & \multirow{2}{*}{vFID$\downarrow$} \\
        \cmidrule(lr){1-3}
        Stage H & Stage A & 3D Skip & & & \\
        \midrule
        $\checkmark$ & $\times$ & $\times$ & 27.89 & 0.083 & 0.206 \\
        $\checkmark$ & $\times$ & $\checkmark$ & 27.50 & 0.099 & 0.249 \\
        $\checkmark$ & $\checkmark$ & $\times$ & \textbf{28.60} & \textbf{0.069} & \textbf{0.178} \\
        \bottomrule
    \end{tabular}
    }
\end{table}

\vspace{-1em}
\paragraph{Vehicle Identity Preservation.}
By tracking the vehicle identity across frames, we can reliably assess temporal consistency.
Fig.~\ref{fig:exp_2} presents a long-range qualitative comparison between R3D2, our method, and the ground truth on distant frames (\eg, Frame 146 and Frame 185). To eliminate domain disparities, R3D2 is retrained on our MirageDrive dataset.

R3D2 often exhibits appearance drift: the inserted vehicle gradually loses consistent shape cues, and its surface reflectance varies noticeably over time. These inconsistencies lead to an unstable vehicle identity across frames. In contrast, Mirage maintains a stable and coherent vehicle identity throughout the sequence. As highlighted in the zoomed-in regions, our method consistently preserves the car’s fine-grained appearance despite long temporal gaps. The resulting appearance remains closely aligned with the ground-truth vehicle, demonstrating the strong long-range temporal consistency achieved by our design.

\subsection{Ablation Study}
\paragraph{VAE Adaptation Strategies.}
In Tab.~\ref{tab:vae_ablation}, we analyze the impact of different VAE adaptation strategies on spatial fidelity and perceptual realism. Using a 2D VAE yields the lowest performance across all metrics. Switching to a 3D VAE brings substantial improvements in PSNR, SSIM, and vFID, demonstrating that modeling video latents in a spatiotemporal space is essential for coherent asset editing.

We further evaluate two hybrid designs. Adding skip connections to the 2D VAE increases PSNR and SSIM, suggesting that direct feature passing is helpful for recovering spatial fidelity. In contrast, injecting 2D encoder features into the 3D decoder (“3D VAE w. 2D Inj.”) yields the best results across all metrics. This design preserves the causal temporal structure of the 3D VAE while restoring the fine spatial details captured by the 2D encoder.

\vspace{-1em}
\paragraph{MirageDrive Ablation.}
Table \ref{tab:dataset_ablation} compares the original R3D2 trained on standard datasets against our version enhanced with MirageDrive trained on our aligned dataset. The original R3D2 achieves a PSNR of 27.61 and vFID of 0.206. By integrating MirageDrive and training on our carefully aligned dataset, we observe significant improvements: PSNR increases by 0.70 to 28.31, indicating superior reconstruction quality. The vFID score decreases from 0.206 to 0.193, demonstrating better distribution alignment with real video data and improved temporal coherence. These results underscore the critical importance of dataset alignment and validate the effectiveness of our MirageDrive.

\vspace{-1em}
\paragraph{Ablation on Training Configurations.}
Tab.~\ref{tab:training_ablation} presents an ablation study evaluating different training configurations. When using only the harmonization training stage (Stage H.), the model achieves a PSNR of 27.89 and vFID of 0.206. The introduction of 3D skip connections leads to performance degradation, due to 3D skip connections breaking the temporal distribution. Our complete framework, which integrates both the harmonization and VAE adaptation stages (Stage H + Stage A), achieves the best performance across all metrics. This demonstrates its ability to produce higher-quality videos with stronger temporal consistency.
\section{Conclusion}
In this paper, we presented Mirage, a one-step video diffusion framework for photorealistic and temporally coherent 3D asset editing in driving scenes. Mirage introduces a temporally-agnostic latent injection mechanism to enhance spatial fidelity in causal 3D VAEs, and a two-stage alignment strategy to provide clean supervision through the MirageDrive dataset. Through extensive experiments, Mirage achieves state-of-the-art realism and stable long-range temporal consistency. We believe Mirage offers a robust foundation for future research on controllable, high-fidelity asset editing for autonomous driving.

{
    \small
    \bibliographystyle{ieeenat_fullname}
    \bibliography{main}
}

\clearpage
\setcounter{page}{1}
\maketitlesupplementary

\noindent This supplementary material is organized as follows:
\begin{itemize}
\item Additional motivation and failure analysis of R3D2 and the R3D3 dataset (Section~\ref{sec:Motivation}).
\item A detailed study of the temporal behavior of our causal 3D VAE (Section~\ref{sec:understanding}).
\item The construction pipeline of MirageDrive and a discussion of the distribution mismatch between scene Gaussians and asset Gaussians (Section~\ref{sec:miragedrive_details}).
\item Qualitative video demos that highlight temporal consistency and controllable editing (Section~\ref{sec:sup6}).
\end{itemize}

\section{Motivation of Our Work}
\label{sec:Motivation}
The motivation of our work arises from two key observations.

First, R3D2 suffers from poor temporal consistency when editing driving videos. As shown in Fig.~\ref{fig:exp_2}, the appearance of the inserted vehicle drifts noticeably across frames. This instability undermines downstream applications that rely on coherent multi-frame behavior, such as long-horizon simulation.

\begin{figure*}[t]
  \centering
   \includegraphics[width=\linewidth]{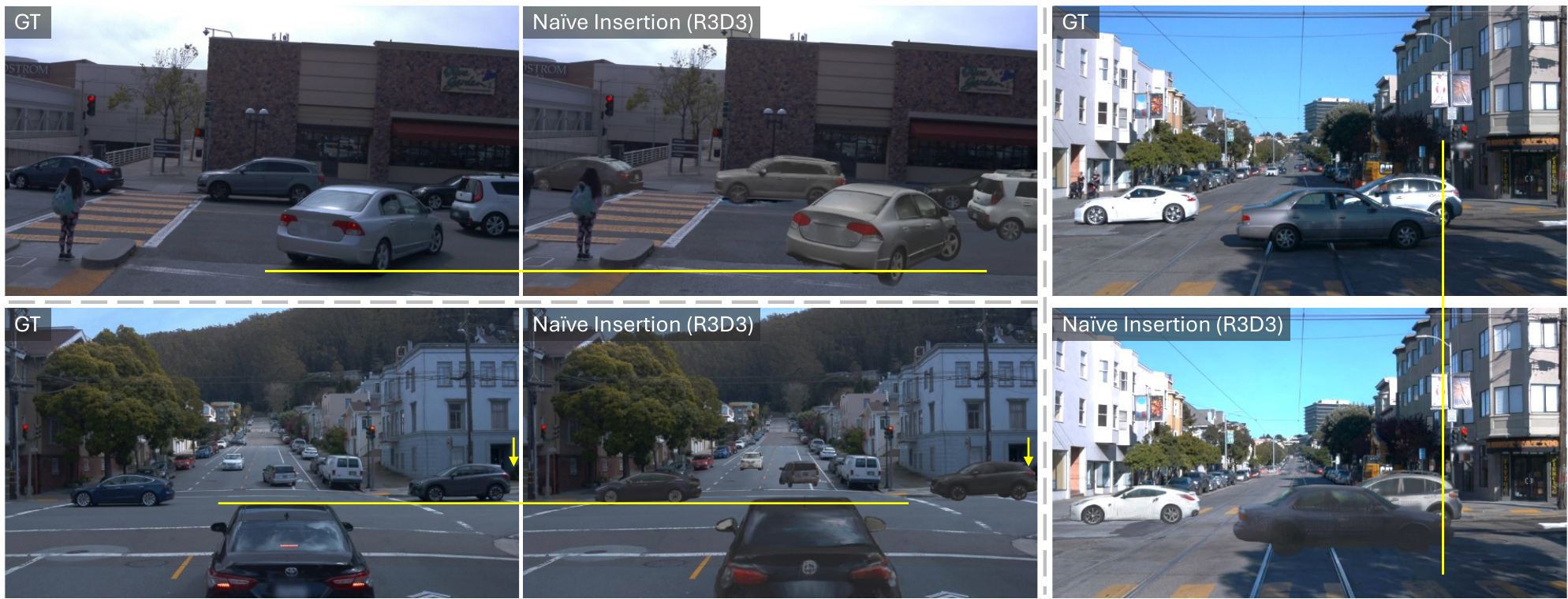}
    \caption{\textbf{Qualitative examples from the R3D3 dataset.}
    Horizontal and vertical reference lines, together with arrows, are used to highlight geometric misalignment between the naïvely inserted assets and the ground-truth objects.
    All frames are taken from the R3D2~\cite{r3d2} paper (Supp.~E).}
   \label{fig:supp_r3d3}
\end{figure*}

Second, the R3D3 dataset itself contains geometric inconsistencies. As illustrated in Fig.~\ref{fig:supp_r3d3}, the highlighted misalignments result in mismatched supervisory pairs during training, which in turn degrades the performance of subsequent harmonization networks.

Together, these issues motivate us to develop a pipeline that ensures both precise spatial alignment and stable temporal behavior, enabling more reliable and photorealistic asset editing in driving scenes.

\section{Understanding the Temporal Behavior of the 3D VAE}
\label{sec:understanding}

\begin{figure}[t]
  \centering
   \includegraphics[width=\linewidth]{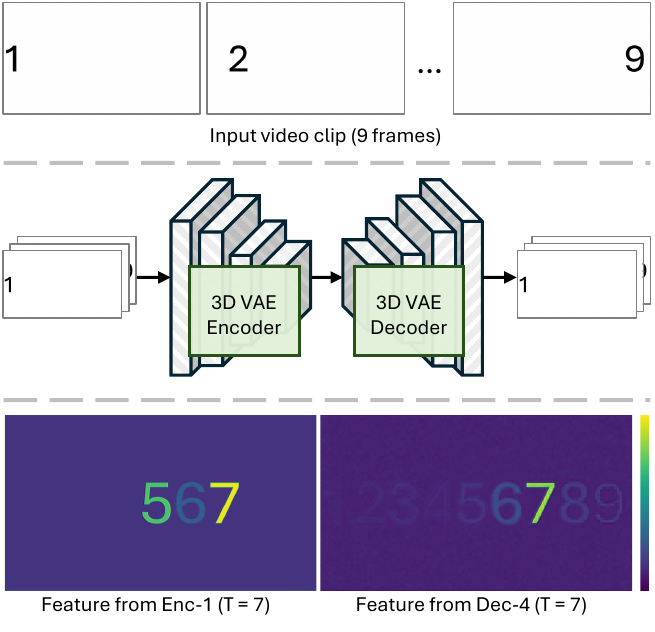}
    \caption{Top: Input 9-frame video clip.
    Middle: Encoding and decoding process of the 3D VAE.
    Bottom: Visualization of feature maps from Encoder Stage~1 and Decoder Stage~4 at $T=7$, obtained by averaging activations across the channel dimension.}
   \label{fig:supp_exp}
\end{figure}

To better understand the temporal behavior of our 3D VAE, we conduct an intuitive experiment using a 9-frame input clip (Fig.~\ref{fig:supp_exp}, top).
Tab.~\ref{tab:3dvae_feature} summarizes the feature shapes across encoder and decoder stages, showing a symmetric hierarchy from Enc-1 to Enc-4 and from Dec-4 to Dec-1.
This structure allows us to examine how temporal information is distributed inside the model.

First, we visualize feature maps from the corresponding encoder and decoder layers.
As shown in Fig.~\ref{fig:supp_exp} (bottom), the activation map at $T=7$ in Enc-1 exhibits strong responses not only at Frame 7 but also at Frame 5.
This observation indicates that encoder features carry mixed temporal support.
In contrast, the corresponding decoder feature (Dec-4 at $T=7$) shows activation concentrated almost exclusively in Frame 7.
The decoder expects content aligned to Frame 7, but the injected encoder features still retain residual signals from Frame 5.
This temporal mismatch leads to leakage and frame-inconsistent artifacts in the reconstructed output.

\begin{table}[t]
\centering
\caption{Feature shapes of the 3D VAE follow the $[C,T,H,W]$ layout with a 9-frame input clip.
The encoder gradually compresses spatial and temporal support from Enc-1 to Enc-4, and the decoder restores them symmetrically from Dec-4 to Dec-1.}
\label{tab:3dvae_feature}
\resizebox{\linewidth}{!}{
    \begin{tabular}{cccc}
    \toprule
    \multicolumn{2}{c}{Encoder} & \multicolumn{2}{c}{Decoder} \\
    \cmidrule(lr){1-2} \cmidrule(lr){3-4}
    Stage & Feature Shapes & Stage & Feature Shapes \\
    \midrule
    Enc-1 & $[128,\,9,\,H,\,W]$     & Dec-4 & $[256,\,9,\,H,\,W]$ \\
    Enc-2 & $[128,\,5,\,H/2,\,W/2]$ & Dec-3 & $[256,\,9,\,H/2,\,W/2]$ \\
    Enc-3 & $[256,\,3,\,H/4,\,W/4]$ & Dec-2 & $[512,\,5,\,H/4,\,W/4]$ \\
    Enc-4 & $[256,\,3,\,H/8,\,W/8]$ & Dec-1 & $[512,\,3,\,H/8,\,W/8]$ \\
    \bottomrule
    \end{tabular}
}
\end{table}

Second, we observe that the temporal axes of the earliest decoder stages (Dec-4 and Dec-3) remain nearly one-to-one with the output frame index.
For example, at $T=7$, the activation at Frame~7 is the dominant response.
These layers primarily reflect frame-aligned spatial content, confirming that they maintain tightly localized temporal support.
Such alignment is crucial for preserving temporal consistency in the final reconstruction.


\section{More Details about MirageDrive}
\label{sec:miragedrive_details}

\begin{figure*}[t]
  \centering
   \includegraphics[width=\linewidth]{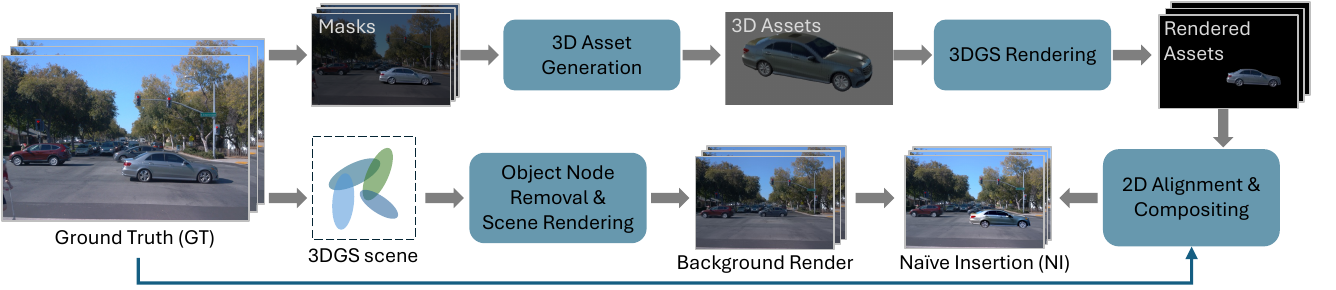}
    \caption{\textbf{MirageDrive data generation pipeline.}
    Starting from ground-truth (GT) driving videos and their 3DGS scene reconstructions, we obtain object masks for the target vehicle and reconstruct a 3D asset.
    We then remove the corresponding object node from the 3DGS scene to render a clean background.
    The rendered 3D asset is finally aligned and composited onto the background to form naïve insertion frames, which provide high-quality NI--GT pairs for training Mirage.}
   \label{fig:supp_data}
\end{figure*}

For each driving scene, we construct MirageDrive in three stages:
scene reconstruction, object-centric asset generation, and background--foreground compositing, as shown in Fig.~\ref{fig:supp_data}.

\subsection{Distribution Mismatch Between Scene Gaussians and Asset Gaussians}
\begin{figure*}[t]
  \centering
   \includegraphics[width=0.75\linewidth]{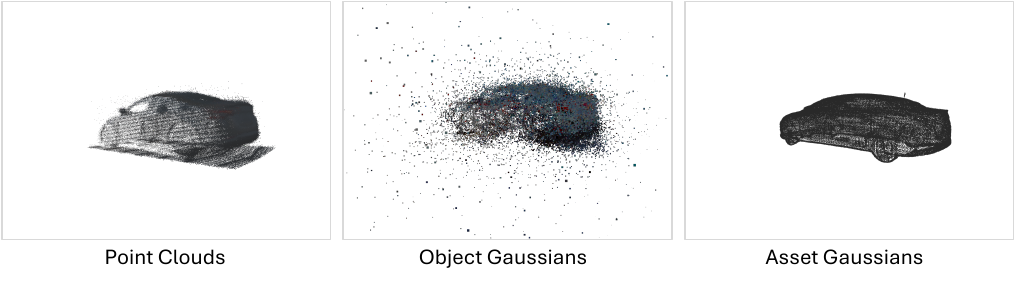}
    \caption{\textbf{Distribution mismatch between scene Gaussians and asset Gaussians.}
    Left: raw point clouds reconstructed from the driving scene.
    Middle: object Gaussians extracted from the 3DGS scene.
    Right: asset Gaussians generated by Trellis~\cite{trellis} exhibit uniformly distributed structures.}
   \label{fig:supp_gaussian}
\end{figure*}

Although scene reconstruction provides object-centric Gaussians, their distributions differ substantially from those of synthesized assets.
As shown in Fig.~\ref{fig:supp_gaussian}, raw point clouds from 3DGS exhibit irregular density patterns.
The extracted object Gaussians remain noisy and highly anisotropic.
In contrast, the generated asset Gaussians produced by Trellis are uniformly distributed. 
This distribution mismatch makes it difficult to directly align the asset with the extracted object Gaussians.
Their centers and scales patterns are not geometrically compatible, which often leads to unstable or inaccurate pose estimation.

To address this issue, MirageDrive adopts a two-stage alignment strategy.
We first perform coarse 3D alignment using camera poses, followed by temporally consistent 2D refinement based on bounding-box edges.
This combination effectively compensates for the distribution disparities and yields stable alignment across time.

\subsection{Scene Reconstruction and Object Selection}
We begin with the ground-truth (GT) driving video and reconstruct its 3D Gaussian Splatting (3DGS) scene using SplatAD~\cite{splatad}.
For each frame, we obtain per-object segmentation masks and temporal tracks of the target vehicle using Grounding-SAM2~\cite{groundingsam}.
Among all tracked instances, we select the one whose segmentation masks remain fully inside the image and exhibit the largest average mask area.
We then crop an image patch around the selected vehicle and use this crop as the reference for asset generation.

\subsection{Asset Generation and Alignment}
Using the object masks, we first identify the actor node in the 3DGS representation and extract all Gaussians associated with this vehicle, forming an object-centric Gaussian subset.
On the asset side, we employ Trellis~\cite{trellis} to generate a high-quality 3D asset from the cropped vehicle image.
We then render this asset using the original camera poses and the poses of the extracted object Gaussians, producing isolated asset renderings that are coarsely aligned in 3D space.

Next, we refine the alignment in the 2D image plane.
We use the per-frame 2D bounding boxes provided by the Waymo Open Dataset~\cite{Waymo} for the original vehicle and adjust the asset placement based on the bounding-box edges over time.
This refinement enforces temporally consistent alignment across all frames.

\subsection{Background Rendering and Data Pair Construction}
On the background side, we remove the corresponding vehicle node from the 3DGS scene and re-render the scene to obtain object-removed frames with a clean background.
We then composite the rendered asset onto these background frames to form naïve insertion (NI) images.
In practice, we perform this process for each candidate asset individually, rather than replacing all vehicles in the scene at once as done in R3D2.

\subsection{Usage at Training and Inference}
The goal of MirageDrive is to construct high-quality NI--GT training pairs that enable our harmonization model to effectively learn spatial and temporal consistency.
During training, MirageDrive provides these paired examples to supervise Mirage.

At inference time, however, our method is not restricted to this specific data construction pipeline.
Any upstream editing system that produces naïve insertion videos can serve as input to Mirage for harmonization.
Furthermore, because the assets are rendered independently of the 3DGS scene, MirageDrive is compatible with a wide range of 3D asset generation pipelines.
In contrast, R3D2 requires dedicated modifications to the scene reconstruction process in order to jointly render both the assets and the background.

\section{Video Demo}
\label{sec:sup6}

We also provide demo videos to showcase the results evaluated in additional scenarios from our test set.
Please refer to the supplementary materials for the \textit{Demo-qU5gAXQ5.mp4} and \textit{Demo-O0ipbuwo.mp4} videos.

\begin{itemize}
\item \textbf{Demo-O0ipbuwo.mp4} compares naïve insertion, R3D2, Mirage, and the ground truth over a frame sequence.
It highlights the superior temporal consistency of Mirage in terms of vehicle identity, appearance stability, and frame-to-frame alignment.

\item \textbf{Demo-qU5gAXQ5.mp4} presents an edited scenario where the front vehicle performs a U-turn, corresponding to the teaser example in the main paper.
This demo illustrates controllable editing and realistic temporal behavior under challenging maneuvers.
\end{itemize}

\end{document}